\pdfoutput=1

\documentclass[11pt]{article}

\usepackage[]{acl}

\usepackage{times}
\usepackage{latexsym}
\usepackage{lipsum}
\usepackage{graphicx}
\usepackage{amsmath}
\usepackage{booktabs}
\usepackage{rotating}
\usepackage{colortbl}
\usepackage{multirow}
\usepackage{listings}
\usepackage{amssymb}
\usepackage{gensymb}
\usepackage{enumitem}

\usepackage[T1]{fontenc}

\usepackage[utf8]{inputenc}

\usepackage{microtype}

\title{A Systematic Survey of \\Text Worlds as Embodied Natural Language Environments}

\author{Peter Jansen \\
  School of Information, University of Arizona, USA \\
  \texttt{pajansen@email.arizona.edu}
  }

\begin{document}
\maketitle
\begin{abstract}
Text Worlds are virtual environments for embodied agents that, unlike 2D or 3D environments, are rendered exclusively using textual descriptions.  These environments offer an alternative to higher-fidelity 3D environments due to their low barrier to entry, providing the ability to study semantics, compositional inference, and other high-level tasks with rich high-level action spaces while controlling for perceptual input. This systematic survey outlines recent developments in tooling, environments, and agent modeling for Text Worlds, while examining recent trends in knowledge graphs, common sense reasoning, transfer learning of Text World performance to higher-fidelity environments, as well as near-term development targets that, once achieved, make Text Worlds an attractive general research paradigm for natural language processing.

\end{abstract}

\section{Introduction}

%
%
\begin{table}[t]
\begin{center}
\scriptsize
\setlength{\tabcolsep}{3pt}
\begin{tabular}{p{0.98\linewidth}} 
\toprule
    \textbf{Zork} \\
\midrule
\textbf{North of House} \\
You are facing the north side of a white house.  There is no door here, and all the windows are barred. \\
\textit{>go north} \\ 
~\\
\textbf{Forest} \\
This is a dimly lit forest, with large trees all around.  One particularly large tree with some low branches stands here. \\
\textit{>climb large tree} \\
~\\
\textbf{Up a Tree} \\
You are about 10 feet above the ground nestled among some large branches. 
On the branch is a small birds nest. In the bird's nest is a large egg encrusted with precious jewels, apparently scavenged somewhere by a childless songbird. \\ 
\textit{>take egg} \\
~\\
Taken. \\
\textit{>climb down tree} \\
~\\
\textbf{Forest} \\
\textit{>go north}   \\
~\\
\textbf{Forest}  \\
This is a dimly lit forest, with large trees all around.  To the east, there appears to be sunlight.    \\
\textit{>go east}    \\
~\\
\textbf{Clearing} \\
You are in a clearing, with a forest surrounding you on the west and south. There is a pile of leaves on the ground.    \\
\textit{>move leaves}    \\
~\\
Done. A grating appears on the ground.    \\
\textit{>open grating}   \\
~\\
The grating is locked.  \\
\bottomrule

\end{tabular}
\caption{\footnotesize An example Text World interactive fiction environment, Zork \cite{lebling1979zork}, frequently used as a benchmark for agent performance. User-entered actions are \textit{italicized}. 
\label{tab:zork-example}}
\end{center}
\vspace{-4mm}
\end{table}

Embodied agents offer an experimental paradigm to study the development and use of semantic representations for a variety of real-world tasks, from household tasks \cite{ALFRED20} to navigation \cite{guss2019minerl} to chemical synthesis \cite{tamari-etal-2021-process}.  While robotic agents are a primary vehicle for studying embodiment \cite[e.g.][]{cangelosi2015developmental}, robotic models are costly to construct, and experiments can be slow or difficult to scale.  Virtual agents and embodied virtual environments help mitigate many of these issues, allowing large-scale simulations to be run in parallel orders of magnitude faster than real world environments \cite[e.g.][]{robothor}, while controlled virtual environments can be constructed for exploring specific tasks -- though this benefit in speed comes at the cost of modeling virtual 3D environments, which can be substantial. 

Text Worlds -- embodied environments rendered linguistically through textual descriptions instead of graphically through pixels (see Table~\ref{tab:zork-example}) -- have emerged as a recent methodological focus that allow studying many embodied research questions while reducing some of the development costs associated with modeling complex and photorealistic 3D environments \cite[e.g.][]{cote2018textworld}.  More than simply reducing development costs, Text Worlds also offer paradigms to study developmental knowledge representation, embodied task learning, and transfer learning at a higher level than perceptually-grounded studies, enabling different research questions that explore these topics in isolation of the open problems of perceptual input, object segmentation, and object classification regularly studied in the vision community
\cite[e.g.][]{he2016deep,szegedy2017inception,zhai2021scaling}.

\subsection{Motivation for this survey}

Text Worlds are rapidly gaining momentum as a research methodology in the natural language processing community. 
Research on agent modeling is regularly reported, while other works aim to standardize tooling, experimental methodologies, and evaluation mechanisms, to rapidly build infrastructure for a sustained research program and community. In spite of this interest, many barriers exist to applying these methodologies, with significant development efforts in the early stages of mitigating those barriers, at least in part.  

In this review, citation graphs of recent articles were iteratively crawled, identifying 108 articles relevant to Text Worlds and other embodied environments that include text as part of the simulation or task.  
Frequent motivations for choosing Text Worlds are highlighted in Section~\ref{sec:why-textworlds}.  Tooling and modeling paradigms (in the form of simulators, intermediate languages, and libraries) are surveyed and compared to higher-fidelity 3D environments in Section~\ref{sec:simulators}, with text environments and common benchmarks implemented with this tooling described in Section~\ref{sec:environments}.  Contemporary focuses in agent modeling, including coupling knowledge graphs, question answering, and common-sense reasoning with reinforcement learning, are identified in Section~\ref{sec:agents}.  Recent contributions to focus areas in world generation, social reasoning, and hybrid text-3D environments are summarized in Section~\ref{sec:contemporary-challenges}, while a distillation of near-term directions for reducing barriers to using Text Worlds more broadly as a research paradigm are presented in Section~\ref{sec:limitations}.

\section{Why use Text Worlds?}
\label{sec:why-textworlds}
For many tasks, Text Worlds can offer advantages over other embodied environment modelling paradigms -- typically in reduced development costs, the ability to model large action spaces, and the ability to study embodied reasoning at a higher level than raw perceptual information. 

{\flushleft\textbf{Embodied Reasoning:}}
Embodied agents have been proposed as a solution to the symbol grounding problem \cite{harnad1990symbol}, or the problem of how concepts acquire real-world meaning.  Humans likely resolve symbol grounding at least partially by assigning semantics to concepts through perceptually-grounded mental simulations \cite{barsalou1999perceptual}.  Using embodied agents that take in perceptual data and perform actions in real or virtual environments offers an avenue for studying semantics and symbol grounding empirically \cite{cangelosi2010integration,bisk2020experience,tamari2020language,tamari2020ecological}.  Text Worlds abstract some of the challenges in perceptual modeling, allowing agents to focus on higher-level semantics, while hybrid worlds that simultaneously render both text and 3D views \cite[e.g.][]{shridhar2020alfworld} help control what kind of knowledge is acquired, and better operationalize the study of symbol grounding. 


{\flushleft\textbf{Ease of Development:}}
Constructing embodied virtual environments typically has steep development costs, but Text Worlds are typically easier to construct for many tasks. Creating new objects does not require the expensive process of creating new 3D models, or performing visual-information-to-object-name segmentation or classification (since the scene is rendered linguistically). 
Similarly, a rich action semantics is possible, and comparatively easy to implement -- while 3D environments typically have one or a small number of action commands \cite[e.g.][]{kolve2017ai2,ALFRED20}, Text Worlds typically implement dozens of action verbs, and thousands of valid \textit{Verb-NounPhrase} action combinations \cite{hausknecht2020interactive}. 


{\flushleft\textbf{Compositional Reasoning:}}
Complex reasoning tasks typically require multi-step (or compositional) reasoning that integrates several pieces of knowledge in an action procedure that arrives at a solution.  In the context of natural language, compositional reasoning is frequently studied through question answering tasks \cite[e.g.][]{yang-etal-2018-hotpotqa,khot2020qasc,xie-etal-2020-worldtree,dalvi2021explaining} or procedural knowledge prediction \cite[e.g.][]{dalvi-etal-2018-tracking,tandon-etal-2018-reasoning,dalvi-etal-2019-everything}. A contemporary challenge is that the number of valid compositional procedures is typically large compared to those that can be tractably annotated as gold, and as such automatically evaluating model performance becomes challenging.
In an embodied environment, an agent's actions have (generally) deterministic consequences for a given environment state, as actions are grounded in an underlying action language \cite[e.g.][]{mcdermott1998pddl} or linear logic \cite[e.g.][]{martens2015ceptre}.
Embodied environments can offer a more formal semantics to study these reasoning tasks, where correctness of novel procedures could be evaluated directly.   

{\flushleft\textbf{Transfer Learning:}}
Training a text-only agent for embodied tasks allows the agent to learn those tasks in a distilled form, at a high-level.  This performance can then be transferred to more realistic 3D environments, where agents pretrained on text versions of the same environment learn to ground their high-level knowledge in low-level perceptual information, and complete tasks faster than when trained jointly \cite{shridhar2020alfworld}.  This offers the possibility of creating simplified text worlds to pretrain agents for challenging 3D tasks that are currently out of reach of embodied agents. 


%
%

%
%
\section{What embodied simulators exist?}
\label{sec:simulators}
Here we explore what simulation engines exist for embodied agents, the trade-offs between high-fidelity (3D) and low-fidelity (text) simulators, and modeling paradigms for text-only environments and agents. 


%
%
\begin{table*}[t]
\centering
{\small
\setlength{\tabcolsep}{3pt}	
\begin{tabular}{cll}  
\specialrule{1pt}{0em}{3pt} 
~     & Example Simulators  &   Typical Characteristics \\
\specialrule{1pt}{3pt}{3pt} 
\multicolumn{3}{c}{3D Environment Simulators}\\
\midrule
\multirow{5}{*}{\includegraphics[scale=0.20]{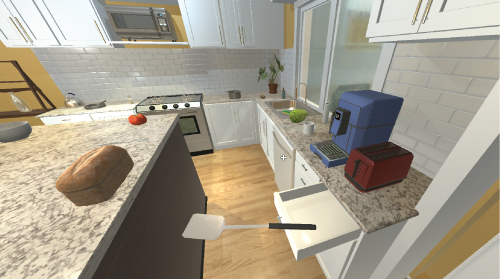}}
        & AI2-Thor \cite{kolve2017ai2}          & $\bullet$ High-resolution 3D visual environments\\
        & CHALET \cite{yan2018chalet}           & $\bullet$ Physics engine for forces, matter, light \\
        & House3D \cite{wu2018building}         & $\bullet$ Adding objects requires complex 3D modeling \\
        & RoboThor \cite{robothor}              & $\bullet$ Limited set of simplified actions \\
        & ALFRED$^{\dagger}$ \cite{ALFRED20}                & $\bullet$ Adding actions is typically expensive \\
        & ALFWorld$^{\dagger}$ \cite{shridhar2020alfworld}  & $\bullet$ Possible transfer to real environments (robotics) \\
\midrule
\multicolumn{3}{c}{Voxel-Based Simulators}\\
\midrule
\multirow{5}{*}{\includegraphics[scale=0.20]{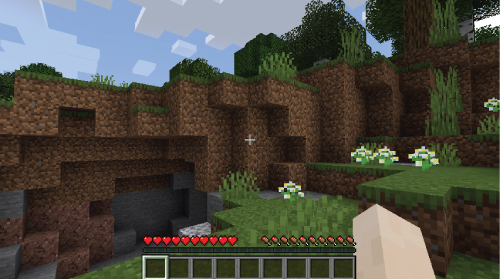}}
        & Malmo \cite{johnson2016malmo}         & $\bullet$ Low-resolution 3D visual environments\\
        & MineRL \cite{guss2019minerl}          & $\bullet$ Simplified physics engine for forces, matter, light\\
        & ~                                     & $\bullet$ Adding objects requires simple 3D modeling \\
        & ~                                     & $\bullet$ Limited set of simplified actions \\
        & ~                                     & $\bullet$ Adding actions is somewhat expensive \\
        & ~                                     & $\bullet$ Limited transfer to real environments \\

\midrule
\multicolumn{3}{c}{Gridworld Simulators}\\
\midrule
\multirow{5}{*}{\includegraphics[scale=0.20]{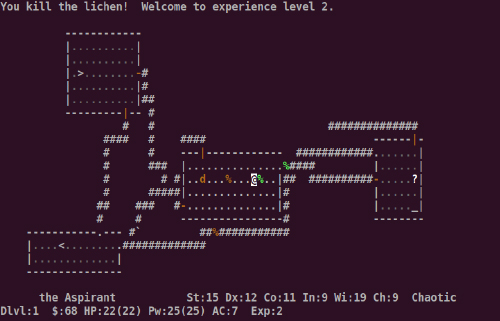}}
        & Rogue-in-a-box \cite{asperti2017rogueinabox}          & $\bullet$ 2D Grid-world rendered graphically/as characters \\
        & BABYAI$^{\dagger}$ \cite{chevalier2018babyai}         & $\bullet$ Low-fidelity physics   \\
        & Nethack LE$^{\dagger}$ \cite{kuttler2020nethack}      & $\bullet$ Adding objects is comparatively easy \\
        & VisualHints$^{\dagger}$ \cite{carta2020visualhints}   & $\bullet$ Small or large action spaces \\
        & Griddly \cite{bamford2021griddly}                     & $\bullet$ Adding object interactions is inexpensive \\
        & ~                                                     & $\bullet$ Limited transfer to real environments \\[1mm]

\midrule
\multicolumn{3}{c}{Text-based Simulators}\\
\midrule
\multirow{5}{*}{\includegraphics[scale=0.20]{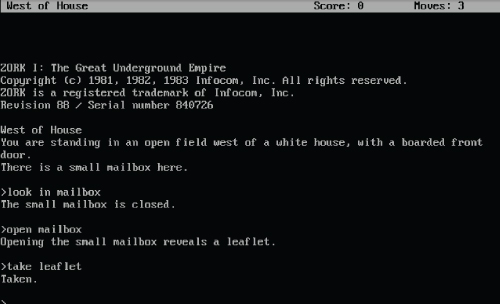}}
        & Z-Machine$^{\dagger}$ \cite{learningZIL1989}          & $\bullet$ Environment described to user using text only \\
        & Inform7$^{\dagger}$ \cite{nelson2006natural}          & $\bullet$ Low-fidelity, conceptual-level physics \\
        & Ceptre$^{\dagger}$ \cite{martens2015ceptre}           & $\bullet$ Adding objects is comparatively easy \\
        & TextWorld$^{\dagger}$ \cite{cote2018textworld}        & $\bullet$ Small or large action spaces \\
        & LIGHT$^{\dagger}$ \cite{urbanek2019learning}          & $\bullet$ Adding actions is comparatively inexpensive \\
        & Jericho$^{\dagger}$ \cite{hausknecht2020interactive}  & $\bullet$ Limited transfer to real environments \\

\specialrule{1pt}{3pt}{0em} 
\end{tabular}
}
\caption{\small Example embodied simulation environments broken down by environment rendering fidelity. $^{\dagger}$ Signifies that the simulator includes a textual component as either input or output to an agent. \label{tab:simulators-examples}}
\end{table*}

\subsection{Simulators}
Simulators provide the infrastructure to implement the environments, objects, characters, and interactions of a virtual world, typically through a combination of a scripting engine to define the behavior of objects and agents, with a rendering engine that provides a view of the world for a given agent or user.
Simulators for embodied agents exist on a fidelity spectrum, from photorealistic 3D environments to worlds described exclusively with language, where a trade-off typically exists between richer rendering and richer action spaces.  This fidelity spectrum (paired with example simulators) is shown in Table~\ref{tab:simulators-examples}, and described briefly below.  Note that many of these higher-fidelity simulators are largely out-of-scope when discussing Text Worlds, except as a means of contrast to text-only worlds, and in the limited context that these simulators make use of text.

{\flushleft\textbf{3D Environment Simulators:}} 
3D simulators provide the user with complex 3D environments, including near-photorealistic environments such as AI2-Thor \cite{kolve2017ai2}, and include physics engines that model forces, liquids, illumination, containment, and other object interactions.  Because of their rendering fidelity, they offer the possibility of inexpensively training robotic models in virtual environments that can then be transferred to the real world \cite[e.g. RoboThor,][]{robothor}.  Adding objects to 3D worlds can be expensive, as this requires 3D modelling expertise that teams may not have.  Similarly, adding agent actions or object-object interactions through a scripting language can be expensive if those actions are outside what is easily implemented in the existing 3D engine or physics engine (like creating \textit{gasses}, or using a pencil or saw to \textit{modify an object}).  Because of this, action spaces tend to be small, and limited to movement, and one (or a small number of) interaction commands.  Some simulators and environments include \textbf{text directives} for an agent to perform, such as an agent being asked to \textit{``slice an apple then cool it''} in the ALFRED environment \cite{ALFRED20}.  Other \textbf{hybrid environments} such as ALFWorld \cite{shridhar2020alfworld} simultaneously render an environment both as 3D as well as in text, allowing agents to learn high-level task knowledge through text interactions, then ground these in environment-specific perceptual input.

{\flushleft\textbf{Voxel-based Simulators:}} 
Voxel-based simulators create worlds from (typically) large 3D blocks, lowering rendering fidelity while greatly reducing the time and skill required to add new objects.  Similarly, creating new agent-object or object-object interactions can be easier because they can generally be implemented in a coarser manner -- though some kinds of basic spatial actions (like rotating an object in increments smaller than 90 degrees) are generally not easily implemented. Malmo \cite{johnson2016malmo} and MineRL \cite{guss2019minerl} offer wrappers and training data to build agents in the popular Minecraft environment.  While the agent's action space is limited in Minecraft (see Table~\ref{tab:example-action-spaces}), the crafting nature of the game (that allows collecting, creating, destroying, or combining objects using one or more voxels) affords exploring a variety of compositional reasoning tasks with a low barrier to entry, while still using a 3D environment.  Text directives, like those in CraftAssist \cite{gray2019craftassist}, allow agents to learn to perform compositional crafting actions in this 3D environment from natural language dialog. 

{\flushleft\textbf{GridWorld Simulators:}} 
2D gridworlds are comparatively easier to construct than 3D environments, and as such more options are available.  GridWorlds share the commonality that they exist on a discretized 2D plane, typically containing a maximum of a few dozen cells on either dimension (or, tens to hundreds of discrete cells, total).  Cells are discrete locations that (in the simplest case) contain up to a single agent or object, while more complex simulators allow cells to contain more than one object, including containers.  Agents move on the plane through simplified spatial dynamics, at a minimum \textit{rotate left, rotate right, and move forward}, allowing the entire world to be explored through a small action space. 

Where gridworlds tend to differ is in their rendering fidelity, and their non-movement action spaces.  In terms of rendering, some \cite[such as BABYAI,][]{chevalier2018babyai} render a world graphically, using pixels, with simplified shapes for improving rendering throughput and reducing RL agent training time.  Others such as NetHack \cite{kuttler2020nethack} are rendered purely as textual characters, owing to their original nature as early terminal-only games.  Some simulators \cite[e.g. Griddly,][]{bamford2021griddly} support a range of rendering fidelities, from sprites (slowest) to shapes to text characters (fastest), depending on how critical rendering fidelity is for experimentation.  As with 3D simulators, hybrid environments \cite[like VisualHints,][]{carta2020visualhints} exist, where environments are simultaneously rendered as a Text World and accompanying GridWorld that provides an explicit spatial map. 

Action spaces vary considerably in GridWorld simulators, owing to the different scripting environments that each affords.  Some environments have a small set of hardcoded environment rules (e.g. BABYAI), while others (e.g. NetHack) offer nearly 100 agent actions, rich crafting, and complex agent-object interactions.  Text can occur in the form of task directives (e.g. \textit{``put a ball next to the blue door''} in BABYAI), partial natural language descriptions of changes in the environmental state (e.g. \textit{``You are being attacked by an orc''} in NetHack), or as full Text World descriptions in hybrid environments.

%
%
\newcommand*{\MyIndent}{\hspace*{0.5cm}}
\begin{table}[t]
\begin{center}
\scriptsize
\setlength{\tabcolsep}{3pt}
\begin{tabular}{l} 
\toprule
    \textbf{Inform7} \\
\midrule
The Kitchen is a room. "A well-stocked Kitchen." \\
The Living Room is north of the Kitchen. \\
A stove is in the Kitchen. A table is in the Kitchen. A plate is on the table.\\
An Apple is on the plate. The Apple is edible. \\
~\\
The cook is a person in the Kitchen. The description is "A busy cook." \\
~\\
The ingredient list is carried by the cook. The description is "A list of \\
\MyIndent ingredients for the cook's favorite recipe.". \\ 
~\\
Instead of listening to the cook:\\ 
\MyIndent say "The cook asks if you can help find some ingredients, and hands \\
\MyIndent \MyIndent you a shopping list from their pocket.";\\
\MyIndent move the ingredient list to the player.\\
~\\
\midrule
    \textbf{Environment Interpreter} \\
\midrule
\textbf{Kitchen}    \\
A well-stocked Kitchen. \\
You can see a stove, a table (on which is a plate (on which is an Apple)) \\
and a cook here.  \\
\textit{>eat apple} \\
~\\
(first taking the Apple)    \\
You eat the Apple. Not bad. \\
\textit{>listen to cook}    \\
~\\
The cook asks if you can help find some ingredients, and hands you a \\
shopping list from their pocket.\\
\bottomrule
\end{tabular}
\caption{\footnotesize An example toy kitchen environment implemented in the Inform7 language \textit{(top)}, and the resulting environment created from those rules \textit{(bottom)}. 
\label{tab:example-inform-code}}
\end{center}
\vspace{-4mm}
\end{table}

{\flushleft\textbf{Text-based Simulators:}} Text World simulators render an agent's world view directly into textual descriptions of their environment, rather than into 2D or 3D graphical renderings.  Similarly, actions the agent wishes to take are typically provided to the simulator as text (e.g. \textit{``read the letter''} in \textit{Zork}), requiring agent models to both parse input text from the environment, and generate output text to to interact with that environment. 

In terms of simulators, the Z-machine \cite{learningZIL1989} is a low-level virtual machine originally designed by Infocom for creating portable interactive fiction novels (such as \textit{Zork}). It was paired with a high-level LISP-like domain-specific language (ZIL) that included libraries for text parsing, and other tools for writing interactive fiction novels.  The Z-machine standard was reverse-engineered by others \cite[e.g.][]{nelson2014zmachine} in an effort to build their own high-level interactive fiction domain-specific languages, and has since become a standard compilation target due to the proliferation of existing tooling and legacy environments.\footnote{A variety of text adventure tooling, including the Adventure Game Toolkit (AGT) and Text Adventure Development System (TADS), was developed starting in the late 1980s, but these simulators have generally not been adopted by the NLP community in favour of the more popular Inform series tools.}

Inform7 \cite{nelson2006natural} is a popular high-level language designed for interactive fiction novels
that allows environment rules to be directly specified in a simplified natural language, substantially lowering the barrier to entry for creating text worlds (see Table~\ref{tab:example-inform-code} for an example).  
The text generation engine allows substantial variation in the way the environments are described, from dry formulaic text to more natural, varied, conversational descriptions. 
Inform7 is compiled to Inform6, an earlier object-oriented scripting language with C-like syntax, which itself is compiled to Z-machine code.

Ceptre \cite{martens2015ceptre} is a linear-logic simulation engine developed with the goal of specifying more generic tooling for operational logics than Inform 7.  TextWorld \cite{cote2018textworld} adapt Ceptre's linear logic state transitions for environment descriptions, and add tooling for generative environments, visualization, and RL agent coupling, all of which is compiled into Inform7 source code.  Parallel to this, the Jericho environment \cite{hausknecht2020interactive} allows inferring relevant vocabulary and template-based object interactions for Z-machine-based interactive fiction games, easing action selection for agents.

\begin{figure}[!t]
	\centering
	\includegraphics[scale=0.35]{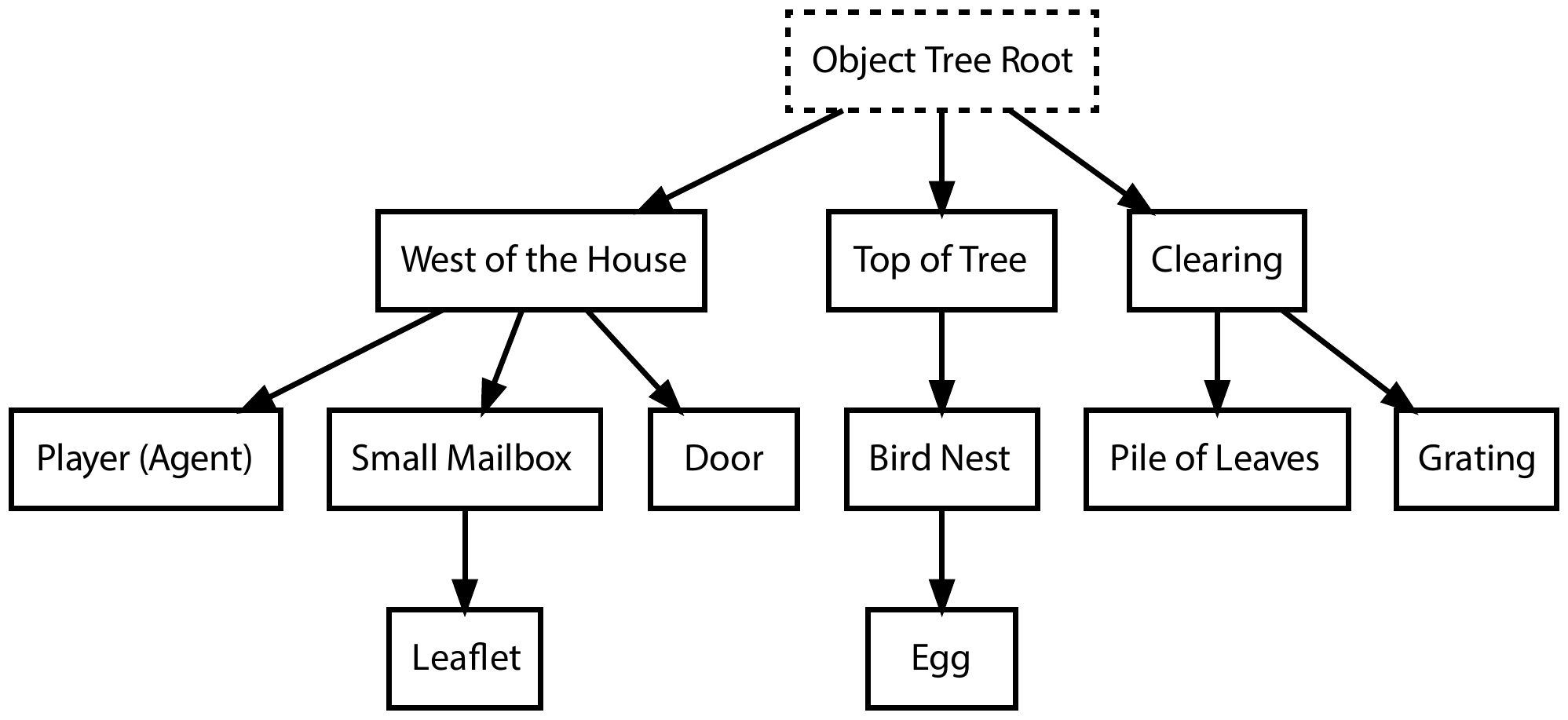}
	\caption{\small An example partial object tree from the interactive fiction game \textit{Zork} \cite{lebling1979zork}.
	\label{fig:object-tree}}
	\vspace{-4mm}
\end{figure}

\subsection{Text World Modeling Paradigms}

\subsubsection{Environment Modelling}
\label{sec:environment-modelling}
Environments are typically modeled as an \textbf{object tree} that stores all the objects in an environment and their nested locations, as well as a set of \textbf{action rules} that implement changes to the objects in the environment based on actions.

{\flushleft\textbf{Objects:}} Because of the body of existing interactive fiction environments for Z-machine environments, and nearly all popular tooling (Inform7, TextWorlds, etc.) ultimately compiling to Z-machine code, object models typically use the Z-machine model \cite{nelson2014zmachine}.  Z-machine objects have \textbf{names} (e.g. ``mailbox''), \textbf{descriptions} (e.g. ``a small wooden mailbox''), binary flags called \textbf{attributes} (e.g. ``is\_container\_open''), and generic \textbf{properties} stored as key-value pairs.  Objects are stored in the \textit{object tree}, which represents the locations of all objects in the environment through parent-child relationships, as shown in Figure~\ref{fig:object-tree}. 

{\flushleft\textbf{Action Rules:}} Action rules describe how objects in the environment change in response to a given world \textbf{state}, which is frequently a collection of preconditions followed by an action taken by an agent (e.g. \textit{``eat the carrot''}), but can also be due to environment states (e.g. a plant dying because it hasn't been watered for a time greater than some threshold). 


Ceptre \cite{martens2015ceptre} and TextWorld \cite{cote2018textworld} use \textbf{linear logic} to represent possible valid \textbf{state transitions}.  In linear logic, a set of \textit{preconditions} in the state history of the world can be consumed by a rule to generate a set of \textit{postconditions}. The set of states are delimited with the linear logic multiplicative conjunction operator ($\otimes$), while preconditions and postconditions are delimited using the linear implication operator ($\multimap$). 
For example, if a player $(P)$ is at location $(L)$, and a container $(C)$ is at that location, and that container is closed, a linear logic rule can be defined to allow the state to be changed to open, as represented by: \vspace{1mm}

\begin{center}
\begin{scriptsize}
\noindent $open(C) :: \$at(P, L) \otimes \$at(C, L) \otimes closed(C) \multimap open(C)$ \\[1mm]
\end{scriptsize}
\end{center}
 
Note that prepending \textit{\$} to a predicate signifies that it will not be consumed by a rule, but carried over to the postconditions.  Here, that means that after opening container $C$, the $P$ and $C$ will still be $at$ location $L$. 

C{\^o}t{\'e} et al.~\shortcite{cote2018textworld} note the limitations in existing implementations of state transition systems for text worlds (such as single-step forward or backward chaining), and suggest future systems may wish to use mature \textbf{action languages} such as STRIPS \cite{fikes1971strips} or GDL \cite{genesereth2005general,thielscher2010general,thielscher2017gdl} as the basis of a world model, though each of these languages have tradeoffs in features (such as object typing) and general expressivity (such as being primarily agent-action centered, rather than easily implementing environment-driven actions) that make certain kinds of modeling more challenging. 
As a proof-of-concept, ALFWorld \cite{shridhar2020alfworld} uses the Planning Domain Definition Language \cite[PDDL,][]{mcdermott1998pddl} to define the semantics for the variety of \textit{pick-and-place} tasks in its text world rendering of the ALFRED benchmark. 


\subsubsection{Agent Modelling}

While environments can be modelled as a collection of states and allowable state transitions (or rules), agents typically have incomplete or inaccurate information about the environment, and must make \textbf{observations} of the environment state through (potentially noisy or inadequate) sensors, and take \textbf{actions} based on those observations.  Because of this, agents are typically modelled as partially-observable Markov decision processes (POMDP) \cite{kaelbling1998planning}, defined by $(S, T, A, R, \Omega, O, \gamma)$. 

A Markov decision process (MDP) contains the state history $(S)$, valid state transitions $(T)$, available actions $(A)$, and (for agent modeling) the expected immediate reward for taking each action $(R)$.  POMDPs extend this to account for partial observability by supplying a finite list of observations the agent can make $(\Omega)$, and an observation function $(O)$ that returns what the agent actually observes from an observation, given the current world state.  For example, the observation function might return \textit{unknown} if the agent tries to examine the contents of a locked container before unlocking it, because the contents cannot yet be observed.  Similarly, when observing the temperature of a cup of tea, the observation function might return coarse measurements \textit{(e.g. hot, warm, cool)} if the agent uses their hand for measurement, or fine-grained measurements \textit{(e.g. $70\degree C$)} if the agent uses a thermometer.  A final discount factor ($\gamma$) influences whether the agent prefers immediate rewards, or eventual (distant) rewards.  The POMDP then serves as a model for a learning framework, typically reinforcement learning (RL), to learn a policy that enables the agent to maximize the reward.  





\section{Text World Environments} 
\label{sec:environments}
%
%
\begin{table*}[t]
\centering
{\footnotesize
\setlength{\tabcolsep}{3pt}	
\begin{tabular}{cll}  
\toprule
Environment     & \# Actions  &   Examples \\
\toprule
\multicolumn{3}{c}{3D Environment Simulators}\\
\midrule
ALFRED          &   7 Command  &   pickup, put, heat, cool, slice, toggle-on, toggle-off                 \\
                &   5 Movement  &   move forward, rotate left, rotate right, look up, look down \\
CHALET          &   1 Command  &   interact    \\
                &   8 Movement  &   move forward, move back, strafe left, strafe right, look left, look right, etc. \\
Malmo/MineRL    &   1 Command  &   attack \\
                &   9 Movement  &   pitch (+/-), yaw (+/-), forward, backward, left, right, jump \\
\midrule
\multicolumn{3}{c}{Gridworld}\\
\midrule
BABYAI          &   4 Command   &   pickup, drop, toggle, done \\
                &   3 Movement  &   turn left, turn right, move forward \\
NETHACK         &   77 Command  &   eat, open, kick, read, tip over, wipe, jump, look, cast, pay, ride, sit, throw, wear, ...\\
                &   16 Movement &   8 compass directions x 2 possibilities (move one step, move far)\\
SocialAI        &   4 Command   &   pickup, drop, toggle, done (from BABYAI)\\
                &   3 Movement  &   turn left, turn right, move forward (from BABYAI)\\
                &   4x16 Text   &   4 templates \textit{(what, where, open, close)} x 16 objects \textit{(exit, bed, fridge)} \\
\midrule
\multicolumn{3}{c}{Text-based}\\
\midrule
ALFWorld        &   11 Command &   goto, take, put, open, close, toggle, heat, cool, clean, inventory, examine  \\
LIGHT           &   11 Command &   get, drop, put, give, steal, wear, remove, eat, drink, hug, hit \\
                &   22 Emotive  &   applaud, blush, cringe, cry, frown, sulk, wave, wink, ...      \\
PEG (Biomedical)    &  35 Command  &   add, transfer, incubate, store, mix, spin, discard, measure, wash, cover, wait, ... \\             
Zork$^{\ddagger}$            &   56 Command &    take, open, read, drop, turn on, turn off, attack, eat, kill, cut, drink, smell, listen, ... \\

\midrule

\specialrule{1pt}{3pt}{0em} 
\end{tabular}
}
\caption{\small Summary statistics for action space complexity for a selection of 3D, gridworld, and text-based environments, demonstrating that Text Worlds generally have large action spaces compared to other simulators. $\ddagger$ Note that while Text World parsers generally recognize a specified number of command verbs, the template extractor in the Jericho framework estimates that common interactive fiction games (including \textit{Zork}) contain between 150 and 300 valid action templates, e.g. \textit{put $<$OBJ$>$ in $<$OBJ$>$} \cite{hausknecht2020interactive}. 
\label{tab:example-action-spaces}}
\end{table*}
Environments are worlds implemented in simulators, that agents explore to perform tasks.  Environments can be simple or complex, test specific or domain-general competencies, be static or generative, and have small or large action spaces compared to higher-fidelity simulators (see Table~\ref{tab:example-action-spaces} for a comparison of action space sizes). 



\subsection{Single Environment Benchmarks}

Single environment benchmarks typically consist of small environments designed to test specific agent competencies, or larger challenging interactive fiction environments that test broad agent competencies to navigate around a diverse world and interact with the environment toward achieving some distant goal. 
Toy environments frequently evaluate an agent's ability to perform compositional reasoning tasks of increasing lengths, such as in the Kitchen Cleanup and related benchmarks \cite{murugesan2020enhancing}. Other toy worlds explore searching environments to locate specific objects \cite{yuan2018counting}, or combining source materials to form new materials \cite{jiang2020wordcraft}. 
While collections of interactive fiction environments are used as benchmarks (see Section~\ref{sec:evironment-collections}), individual environments frequently form single benchmarks.  Zork \cite{lebling1979zork} and its subquests are medium-difficulty environments frequently used in this capacity, while Anchorhead \cite{anchorhead} is a hard-difficulty environment where state-of-the-art performance remains below 1\%. 

\subsection{Domain-specific Environments}

Domain-specific environments allow agents to learn highly specific competencies relevant to a single domain, like science or medicine, while typically involving more modeling depth than toy environments. 
Tamari et al.~\shortcite{tamari-etal-2021-process} create a TextWorld environment for wet lab chemistry protocols, that describe detailed step-by-step instructions for replicating chemistry experiments.  These text-based simulations can then be represented as \textit{process execution graphs (PEG)}, which can then be run on real lab equipment.  A similar environment exists for the materials science domain \cite{tamari2019playing}.  

\subsection{Environment Collections as Benchmarks}
\label{sec:evironment-collections}
To test the generality of agents, large collections of interactive fiction games (rather than single environments) are frequently used as benchmarks.  While the Text-Based Adventure AI Shared Task initially evaluated on a single benchmark environment, later instances switched to evaluating on 20 varied environments to gauge generalization \cite{atkinson2019text}. Fulda et al.~\shortcite{fulda2017can} created a list of 50 interactive fiction games to serve as a benchmark for agents to learn common-sense reasoning.  C{\^o}t{\'e} et al.~\shortcite{cote2018textworld} further curate this list, replacing 20 games without scores to those more useful for RL agents.  The Jericho benchmark \cite{hausknecht2020interactive} includes 32 interactive fiction games that support Jericho's in-built methods for score and world-change detection, out of a total of 56 games known to support these features.

\subsection{Generative Environments}

A difficulty with statically-initialized environments is that because their structure is identical each time the simulation is run, rather than learning general skills, agents quickly overfit to learn solutions to an exact instantiation of a particular task and environment, and rarely generalize to unseen environments \cite{chaudhury-etal-2020-bootstrapped}.  Procedurally generated environments help address this need by generating variations of environments centered around specific goal conditions. 

The TextWorld simulator \cite{cote2018textworld} allows specifying high-level parameters such as the number of rooms, objects, and winning conditions, then uses a random walk to procedurally generate environment maps in the Inform7 language meeting those specifications, using either forward or backward chaining during generation to verify tasks can be successfully completed in the random environment.  As an example, the First TextWorld Problems shared task\footnote{\url{https://competitions.codalab.org/competitions/21557}} used TextWorld to generate 5k variations of a cooking environment, divided into train, development, and test sets.  Similarly, Murugesan et al.~\shortcite{murugesan2020text} introduce TextWorld CommonSense (TWC), a simple generative environment for household cleaning tasks, modelled as a \textit{pick-and-place} task where agents must pick up common objects from the floor, and place them in their common household locations (such as placing \textit{shoes} in a \textit{shoe cabinet}).  Other related environments include Coin Collector \cite{yuan2018counting}, a generative environment for a navigation task, and Yin et al.'s ~\shortcite{yin2019learn} procedurally generated environment for cooking tasks. 

Adhikari et al.~\shortcite{adhikari2020learning} generate a large set of recipe-based cooking games, where an agent must precisely follow a cooking recipe that requires collecting tools (e.g. \textit{a knife}) and ingredients (e.g. \textit{carrots}), and processing those ingredients correctly (e.g. \textit{dice carrots, cook carrots}) in the correct order. Jain et al.~\shortcite{jain2020algorithmic} propose a similar synthetic benchmark for multi-step compositional reasoning called SaladWorld. 
In the context of question answering, Yuan et al.~\shortcite{yuan-etal-2019-interactive} procedurally generate a simple environment that requires an agent to search and investigate attributes of objects, such as verifying their existence, locations, or specific attributes (like edibility). 
On the balance, while tooling exists to generate simple procedural environments, when compared to classic interactive fiction games (such as \textit{Zork}), the current state-of-the-art allows for generating only relatively simple environments with comparatively simple tasks and near-term goals than human-authored interactive fiction games. 


%
%
\begin{table*}[!t]
\centering
\footnotesize
\vspace{3mm}
\begin{tabular}{lccccccccc}
Model &
\begin{rotate}{60} Detective (E) \end{rotate} &
\begin{rotate}{60} Zork1 (M) \end{rotate} &
\begin{rotate}{60} Zork3 (M) \end{rotate} &
\begin{rotate}{60} OmniQuest (M) \end{rotate} &
\begin{rotate}{60} Spirit (H) \end{rotate} &
\begin{rotate}{60} Enchanter (H) \end{rotate} \\[1mm] \hline 
\vspace{-2mm}
~\\[1pt]

DRRN \cite{he-etal-2016-deep-reinforcement} &   0.55    &   0.09    &   0.07    &   0.20    &   0.05    &   0.00    &       \\[1mm]  
BYU-Agent \cite{fulda2017can}           &   0.59    &   0.03    &   0.00    &   0.10    &   0.00    &   0.01    &       \\[1mm]  
Golovin \cite{kostka2017text}           &   0.20    &   0.04    &   0.10    &   0.15    &   0.00    &   0.01    &       \\[1mm]  
AE-DQN \cite{zahavy2018learn}           &   --      &   0.05    &   --      &   --      &   --      &   --      &       \\[1mm]  
NeuroAgent \cite{rajalingam2019neuroevolution}  &   0.19    &   0.03    &   0.00    &   0.20    &   0.00    &   0.00    &       \\[1mm]  
NAIL \cite{hausknecht2019nail}          &   0.38    &   0.03    &   0.26    &   --      &   0.00    &   0.00    &       \\[1mm]  
CNN-DQN \cite{yin2019comprehensible}    &   --      &   0.11    &   --      &   --      &   --      &   --      &       \\[1mm]  
IK-OMP \cite{tessler2019action}         &   --      &   1.00    &   --      &   --      &   --      &   --      &       \\[1mm]  
TDQN \cite{hausknecht2020interactive}   &   0.47    &   0.03    &   0.00    &   0.34    &   0.02    &   0.00    &       \\[1mm]
KG-A2C \cite{ammanabrolu2020graph}      &   0.58    &   0.10    &   0.01    &   0.06    &   0.03    &   0.01    &       \\[1mm]  
SC \cite{jain2020algorithmic}           &   --      &   0.10    &   --      &   --      &   0.0     &   --      &       \\[1mm]  
CALM (N-gram) \cite{yao-etal-2020-keep} &   0.79    &   0.07    &   0.00    &   0.09    &   0.00    &   0.00    &       \\[1mm]  
CALM (GPT-2) \cite{yao-etal-2020-keep}  &   0.80    &   0.09    &   0.07    &   0.14    &   0.05    &   0.01    &       \\[1mm]
RC-DQN \cite{guo2020interactive}        &   0.81    &   0.11    &   0.40    &   0.20    &   0.05    &   0.02    &       \\[1mm]  
MPRC-DQN \cite{guo2020interactive}      &   0.88    &   0.11    &   0.52    &   0.20    &   0.05    &   0.02    &       \\[1mm]  
SHA-KG \cite{xu2020deep}                &   0.86    &   0.10    &   0.10    &   --      &   0.05    &   0.02    &       \\[1mm]  
MC!Q*BERT \cite{ammanabrolu2020avoid}   &   0.92    &   0.12    &   --      &   --      &   0.00    &   --      &       \\[1mm]
INV-DY \cite{yao2021reading}            &   0.81    &   0.12    &   0.06    &   0.11    &   0.05    &   --      &       \\[1mm]

\end{tabular}
\caption{\label{tab:agent-performance} \footnotesize Agent performance on benchmark interactive fiction environments.  All performance values are normalized to maximum achievable scores in a given environment. Due to the lack of standard reporting practice, performance reflects values reported for agents, but is unable to hold other elements (such as number of training epochs, number of testing epochs, reporting average vs maximum performance) constant. Parentheses denote environment difficulty (E:Easy, M:Medium, H:Hard) as determined by the Jericho benchmark \cite{hausknecht2020interactive}. }
\end{table*}

%
%
\section{Text World Agents}
\label{sec:agents}
Recently a large number of agents have been proposed for Text World environments.
This section briefly surveys common modeling methods, paradigms, and trends, with the performance of recent agents on common easy, medium, and hard interactive fiction games \cite[as categorized by the Jericho benchmark, ][]{hausknecht2020interactive} shown in Table~\ref{tab:agent-performance}.




{\flushleft\textbf{Reinforcement Learning:}} 
While some agents rely on learning frameworks heavily coupled with heuristics \cite[e.g.,][Golovin]{kostka2017text}, owing to the sampling benefits afforded by operating in a virtual environment, the predominant modeling paradigm for most contemporary text world agents is reinforcement learning. 
Narasimhan et al.~\shortcite{narasimhan-etal-2015-language} demonstrate that ``Deep-Q Networks'' (DQN) \cite{mnih2015human} developed for Atari games can be augmented with LSTMs for representation learning in Text Worlds, which outperform simpler methods using n-gram bag-of-words representations. 
He et al.~\shortcite[DRRN]{he-etal-2016-deep} extend this to build the Deep Reinforcement Relevance Network (DRRN), an architecture that uses separate embeddings for the state space and actions, to improve both training time and performance. 
Madotto et al.~\shortcite{madotto2020exploration} show that the Go-Explore algorithm \cite{ecoffet2019go}, which periodically returns to promising but underexplored areas of a world, can achieve higher scores than the DRRN with fewer steps. 
Zahvey et al.~\shortcite[AE-DQN]{zahavy2018learn} use an Action Elimination Network (AEN) to remove sub-optimal actions, showing improved performance over a DQN on \textit{Zork}. 
Yao et al~\shortcite[CALM]{yao-etal-2020-keep} use a GPT-2 language model trained on human gameplay to reduce the space of possible input command sequences, and produce a shortlist of candidate actions for an RL agent to select from. 
Yao et al.~\shortcite[INV-DY]{yao2021reading} demonstrate that semantic modeling is important, showing that models that either encode semantics through an inverse dynamic decoder, or discard semantics by replacing words with unique hashes, have different performance distributions in different environments.  
Taking a different approach, Tessler et al.~\shortcite[IK-OMP]{tessler2019action} show that imitation learning combined with a compressed sensing framework can solve Zork when restricted to a vocabulary of 112 words extracted from walk-through examples. 

{\flushleft\textbf{Constructing Graphs:}} 
Augmenting reinforcement learning models to produce knowledge graphs of their beliefs can reduce training time and improve overall agent performance \cite{ammanabrolu-riedl-2019-playing}. 
Ammanabrolu et al.~\shortcite[KG-A2C]{ammanabrolu2020graph} demonstrate a method for training an RL agent that uses a knowledge graph to model its state-space, and use a template-based action space to achieve strong performance across a variety of interactive fiction benchmarks. 
Adhikari et al.~\shortcite{adhikari2020learning} demonstrate that a Graph Aided Transformer Agent (GATA) is able to learn implicit belief networks about its environment, improving agent performance in a cooking environment. 
Xu et al.~\shortcite[SHA-KG]{xu2020deep} extend KG-A2C to use use hierarchical RL to reason over subgraphs, showing substantially improved performance on a variety of benchmarks. 

To support these modelling paradigms, Zelinka et al.~\shortcite{zelinka2019building} introduce TextWorld KG, a dataset for learning the subtask of updating knowledge graphs based on text world descriptions in a cooking domain, and show their best ensemble model is able to achieve 70 F1 at this subtask. 
Similarly, Annamabrolu et al.~\shortcite{ammanabrolu2021modeling} introduce JerichoWorld, a similar dataset for world modeling using knowledge graphs but on a broader set of interactive fiction games, and subsequently introduce WorldFormer \cite{ammanabrolu2021learning}, a multi-task transformer model that performs well at both knowledge-graph prediction and next-action prediction tasks.

{\flushleft\textbf{Question Answering:}} 
Agents can reframe Text World tasks as question answering tasks to gain relevant knowledge for action selection, with these agents providing current state-of-the-art performance across a variety of benchmarks.
Guo et al.~\shortcite[MPRC-DQN]{guo-etal-2020-interactive} use multi-paragraph reading comprehension (MPRC) techniques to ask questions that populate action templates for agents, substantially reducing the number of training examples required for RL agents while achieving strong performance on the Jericho benchmark. 
Similarly, Ammanabrolu et al.~\shortcite[MC!Q*BERT]{ammanabrolu2020avoid} use contextually-relevant questions (such as \textit{``Where am I?'', ``Why am I here?''}, and \textit{``What do I have?''}) to populate their knowledge base to support task completion. 

{\flushleft\textbf{Common-sense Reasoning:}} 
Agents arguably require a large background of common-sense or world knowledge to perform embodied reasoning in general environments. 
Fulda et al.~\shortcite{fulda2017can} extract common-sense affordances from word vectors trained on Wikipedia using word2vec \cite{mikolov2013distributed}, and use this to increase performance on interactive fiction games, as well as (more generally) on robotic learning tasks \cite{fulda2017harvesting}. 
Murugesan et al.~\shortcite{murugesan2020enhancing} combine the ConceptNet \cite{speer2017conceptnet} common-sense knowledge graph with an RL agent that segments knowledge between general world knowledge, and specific beliefs about the current environment, demonstrating improved performance in a cooking environment. 
Similarly, Dambekodi et al.~\shortcite{dambekodi2020playing} demonstrate that RL agents augmented with either COMET \cite{bosselut2019comet}, a transformer trained on common-sense knowledge bases, or BERT \cite{devlin-etal-2019-bert}, which is hypothesized to contain common-sense knowledge, outperform agents without this knowledge on the interactive fiction game \textit{9:05}. 
In the context of social reasoning, Ammanabrolu et al.~\shortcite{ammanabrolu-etal-2021-motivate} create a fantasy-themed knowledge graph, ATOMIC-LIGHT, and show that an RL agent using this knowledge base performs well at the social reasoning tasks in the LIGHT environment. 

\subsection{Generalization across environments}
\label{sec:generalization-across-environments}
Agents trained in one environment rarely transfer their performance to other environments. 
Yuan et al.~\shortcite{yuan2018counting} propose that dividing learning into segmented episodes can improve transfer learning to unseen environments, and demonstrate transfer to more difficult procedurally-generated versions of Coin Collector environments. Yin et al.~\shortcite{yin2019learn} demonstrate that separating knowledge into universal (environment-general) and instance (environment-specific) knowledge using curriculum learning improves model generalization. 

Ansari et al.~\shortcite{ansari2018language} suggest policy distillation may reduce overfitting in Text Worlds, informed by experiments using an LSTM-DQN model on 5 toy environments designed to test generalization.  Similarly, Chaudrey et al.~\shortcite{chaudhury-etal-2020-bootstrapped} demonstrate that policy distillation can be used on RL agents to reduce overfitting and improve generalization to unseen environments using 10x fewer training episodes when training on the Coin Collector environment, and evaluating in a cooking environment. Adolphs et al.~\shortcite{adolphs2020ledeepchef} combine an actor-critic approach with hierarchical RL to demonstrate agent generalization on cooking environments. 

Yin et al.~\shortcite{yin2020zero} propose a method for factorizing Q values that allows agents to better learn in multi-task environments where tasks have different times to reaching rewards.  They empirically demonstrate this novel Q-factorization method requires an order of magnitude less training data, while enabling zero-shot performance on unseen environments.  A t-SNE plot shows that Q-factorization produces qualitatively different and well-clustered learning of game states compared to conventional Q learning.


\section{Contemporary Focus Areas}
\label{sec:contemporary-challenges}

\subsection{World Generation}
Generating detailed environments with complex tasks is labourious, while randomly generating environments currently provides limited task complexity and environment cohesiveness.  World generation aims to support the generation of complex, coherent environments, either through better tooling for human authors \cite[e.g.][]{temprado2019online}, or automated generation systems that may or may not have a human-in-the-loop. 
Fan et al.~\shortcite{fan2020generating} explore creating cohesive game worlds in the LIGHT environment using a variety of embedding models including Starspace \cite{wu2018starspace} and BERT \cite{devlin-etal-2019-bert}.  Automatic evaluations show performance of between 36-47\% in world building, defined as cohesively populating an environment with locations, objects, and characters.  Similarly, human evaluation shows that users prefer Starspace-generated environments over those generated by a random baseline.  In a more restricted domain, Ammanabrolu et al.~\shortcite{ammanabrolu2019toward} show that two models, one Markov chain model, the other a generative language model (GPT-2), are capable of generating quests in a cooking environment, while there is a tradeoff between human ratings of quest creativity and coherence.

Ammanabrolu et al.~\shortcite{ammanabrolu2020bringing} propose a large-scale end-to-end solution to world generation that automatically constructs interactive fiction environments based on a story (such as \textit{Sherlock Holmes}) provided as input.  Their system first builds a knowledge graph of the story by framing KG construction as a question answering task, using their model (AskBERT) to populate this graph.  The system then uses either a rule-based baseline or a generative model (GPT-2) to generate textual descriptions of the world from this knowledge graph.  User studies show that humans generally prefer these neural-generated worlds to the rule-generated worlds (measured in terms of interest, coherence, and genre-resemblance), but that neural-generated performance still substantially lags behind that of human-generated worlds.


\subsection{Social Modeling and Dialog}
While embodied environments provide an agent the opportunity to explore and interact with a physical environment, they also provide an opportunity for exploring human-to-agent or agent-to-agent social interaction grounded in particular situations, consisting of collections of locations, and the objects and agents in those locations. 

The LIGHT platform \cite{urbanek-etal-2019-learning} is large text-only dataset designed specifically to study grounded social interaction.  More than modeling environments and physical actions, LIGHT includes a large set of 11k crowdsourced situated agent-to-agent dialogs, and sets of emotive as well as physical actions.  Specifically, LIGHT emulates a multi-user dungeon (MUD) dialog scenario, and includes 663 interconnected locations, which can be populated with 1755 characters and 3462 objects.  In addition to 11 physical actions (\textit{get, eat, wear, etc.}), it includes 22 emotive actions (\textit{e.g. applaud, cry, wave}) that affect agent behavior.  At each turn, an agent can say something (in natural language) to another agent, take a physical action, and take an emotive action.  Urbanek et al.~\shortcite{urbanek-etal-2019-learning} propose a next-action prediction task (modelled as picking the correct dialog, physical action, and emotive action from 20 possible choices for each), and evaluate both model (BERT) and human performance.  Human vs model performance reaches 92 vs 71 for predicting dialog, 72 vs 52 for physical actions, and 34 vs 29 for emotives, demonstrating the difficulty of the task for models, and the large variance in predicting accompanying emotives for humans.  Others have proposed alternate models, such as Qi et al.~\shortcite{qiu2021towards}, who develop a mental state parser that explicitly models an agent's mental state (including both the agent's physical observations and beliefs) in a graphical formalism to increase task performance.

As an alternate task, Prabhumoye et al.~\shortcite{prabhumoye2020love} use LIGHT to explore persuasion or task-cueing, requiring the player to say something that would cause an agent to perform a specific situated action (like \textit{picking up a weapon} while in the armory) or emotive (like \textit{hugging} the player).  This is challenging, with the best RL models succeeding only about half the time.  

Kova{\v{c} et al.~\shortcite{kovavc2021socialai} propose using pragmatic frames \cite{vollmer2016pragmatic} as a means of implementing structured dialog sessions between agents (with specific roles) in embodied environments.  To explore this formalism, they extend BABYAI to include multiple non-player character (NPC) agents that the player must seek information from to solve their \textit{SocialAI} navigation task.  Inter-agent communication uses 4 frames and 16 referents, for a total of 64 possible utterances, and requires modeling belief states as NPCs may be deceptive.  An RL agent performs poorly at this task, highlighting the difficulty of modeling even modest social interactions that involve communicating with multiple agents. 


As environments move from toy problems to large spaces that contain multiple instances of the same category of object (like more than one cup or house), agents that communicate (e.g. \textit{``pick up the cup'', ``go to the house''}) have to resolve which \textit{cup} or \textit{house} is referenced. Kelleher et al.~\shortcite{kelleher2019referring} propose resolving ambiguous references in embodied agent dialog by using a decaying working memory, similar to models of human short-term memory \cite[e.g.][]{atkinson1968human}, that resolve to the most recently observed or recently thought-about object.


\subsection{Hybrid 3D-Text Environments}

Hybrid simulators that can simultaneously render worlds both graphically (2D or 3D) as well as textually offer a mechanism to quickly learn high-level tasks without having to first solve grounding or perceptual learning challenges. 
The ALFWorld simulator \cite{shridhar2020alfworld} combines the ALFRED 3D home environment \cite{ALFRED20} with a simultaneous TextWorld interface to that same environment, and introduce the BUTLER agent, which shows increased task generalization on the 3D environment when first trained on the text world.
Prior to ALFWorld, Jansen \shortcite{jansen2020visually} showed that a language model (GPT-2) was able to successfully generate detailed step-by-step textual descriptions of ALFRED task trajectories for up to 58\% of unseen cases using task descriptions alone, without visual input.
Building on this, Micheli~\shortcite{micheli2021language} confirmed GPT-2 also performs well on the text world rendering of ALFWorld, and is able to successfully complete goals in 95\% of unseen cases.  
Taken together, these results show the promise of quickly learning complex tasks at a high-level in a text-only environment, then transferring this performance to agents grounded in more complex environments.

\section{Contemporary Limitations and Challenges}
\label{sec:limitations}

{{\flushleft\textbf{Environment complexity is limited, and it's currently difficult to author complex worlds.}}  Two competing needs are currently at odds: the desire for complex environments to learn complex skills, and the desire for environment variation to encourage robustness in models.  Current tooling emphasizes creating varied procedural environments, but those environments have limited complexity, and require agents to complete straightforward tasks.  Economically creating complex, interactive environments that simulate a significant fraction of real world interactions are still well beyond current simulators or libraries -- but required for higher-fidelity interactive worlds that have multiple meaningful paths toward achieving task goals.  Generating these environments semi-automatically \cite[e.g.][]{ammanabrolu2020bringing} may offer a partial solution.  Independent of tooling, libraries and other middleware offer near-term solutions to more complex environment modeling, much in the same way 3D game engines are regularly coupled with physics engine middleware to dramatically reduce the time required to implement forces, collisions, lighting, and other physics-based modeling.  Currently, few analogs exist for text worlds.  The addition of a \textit{chemistry engine} that knows that ice warmed above the freezing point will change to liquid water, or a \textit{generator engine} that knows the sun is a source of sunlight during sunny days, or an \textit{observation engine} that knows tools (like microscopes or thermometers) can change the observation model of a POMPD -- may offer tractability in the form of modularization.  Efforts using large-scale crowdsourcing to construct knowledge bases of common-sense knowledge \cite[e.g., ATOMIC,][]{sap2019atomic} may be required to support these efforts.

{{\flushleft\textbf{Current planning languages offer a partial solution for environment modelling.}} While simulators partially implement facilities for world modeling, some \cite[e.g.][]{cote2018textworld,shridhar2020alfworld} suggest using mature planning languages like STRIPS \cite{fikes1971strips} or PDDL \cite{mcdermott1998pddl} for more full-featured modeling.  This would not be without significant development effort -- existing implementations of planning languages typically assume full-world observability (in conflict with POMPD modelling), and primarily agent-directed state-space changes, making complex world modeling with partial observability, and complex environment processes (such as plants that require water and light to survive, or a sun that rises and sets causing different items to be observable in day versus night) outside the space of being easily implemented with off-the-shelf solutions.  In the near-term, it is likely that a domain-specific language specific to complex text world modeling would be required to address these needs while simultaneously reducing the time investment and barrier-to-entry for end users.

{{\flushleft\textbf{Analyses of environment complexity can inform agent design and evaluation.}}  Text world articles frequently emphasize agent modeling contributions over environment, methodological, or analysis contributions -- but these contributions are critical, especially in the early stages of this subfield.  Agent performance in easy environments has increased incrementally, while medium-to-hard environments have seen comparatively modest improvements.  Agent performance is typically reported as a distribution over a large number of environments, and the methodological groundwork required to understand when different models exceed others in time or performance over these environment distributions is critical to making forward progress.  Transfer learning in the form of training on one set of environments and testing on others has become a standard feature of benchmarks \cite[e.g.][]{hausknecht2020interactive}, but focused contributions that work to precisely characterize the limits of what can be learned from (for example) \textit{OmniQuest} and transferred to \textit{Zork}, and what capacities must be learned elsewhere, will help inform research programs in agent modeling and environment design.

{{\flushleft\textbf{Transfer learning between text world and 3D environments.}} Tasks learned at a high-level in text worlds help speed learning when those same models are transferred to more complex 3D environments \cite{shridhar2020alfworld}.  This framing of transfer learning may in some ways resemble how humans can converse about plans for future actions in locations remote from those eventual actions (like classrooms).  As such, text-plus-3D environment rendering shows promise as a manner of controlling for different sources of complexity in input to multi-modal task learning (from high-level task-specific knowledge to low-level perceptual knowledge), and appears a promising research methodology for future work.  
\bibliography{anthology,custom}
\bibliographystyle{acl_natbib}

%

\end{document}